\documentclass{article}

\usepackage{arxiv}

\usepackage[utf8]{inputenc} 
\usepackage[T1]{fontenc}    
\usepackage{hyperref}       
\usepackage{url}            
\usepackage{booktabs}       
\usepackage{amsfonts}       
\usepackage{nicefrac}       
\usepackage{microtype}      
\usepackage{lipsum}
\usepackage{graphicx}

\usepackage{graphicx}
\usepackage{multirow}
\usepackage{listings,multicol}
\usepackage{algpseudocode}
\usepackage{lipsum}
\usepackage{fancyvrb}
\usepackage{parcolumns}
\usepackage{verbatim}

\usepackage[T1]{fontenc}
\usepackage{beramono}
\usepackage{xcolor}
\usepackage[numbers,sort&compress]{natbib}

\definecolor{dkgreen}{rgb}{0,0.6,0}
\definecolor{gray}{rgb}{0.5,0.5,0.5}
\definecolor{mauve}{rgb}{0.58,0,0.82}

\usepackage{amsmath}

\lstdefinestyle{myScalastyle2}{
  frame=tb,
  float=*,
  language=scala,
  aboveskip=3mm,
  belowskip=3mm,
  showstringspaces=false,
  columns=flexible,
  basicstyle={\small\ttfamily},
  numbers=none,
  numberstyle=\tiny\color{gray},
  keywordstyle=\color{blue},
  commentstyle=\color{dkgreen},
  stringstyle=\color{mauve},
  frame=single,
  breaklines=true,
  breakatwhitespace=true,
  tabsize=3,
}

\lstdefinestyle{myScalastyle}{
  frame=tb,
  language=scala,
  aboveskip=3mm,
  belowskip=3mm,
  showstringspaces=false,
  columns=flexible,
  basicstyle={\small\ttfamily},
  numbers=none,
  numberstyle=\tiny\color{gray},
  keywordstyle=\color{blue},
  commentstyle=\color{dkgreen},
  stringstyle=\color{mauve},
  frame=single,
  breaklines=true,
  breakatwhitespace=true,
  tabsize=3,
}

\title{Large-scale text processing pipeline with Apache Spark}

\author{
  Alexey Svyatkovskiy \\
  Princeton University\\
  Princeton, NJ 08540  \\
   \And
  Kosuke Imai \\
  Princeton University\\
  Princeton, NJ 08540 \\
   \And
  Mary Kroeger \\
  Princeton University\\
  Princeton, NJ 08540 \\
  \And
  Yuki Shiraito \\
  Princeton University\\
  Princeton, NJ 08540 \\
}

\begin{document}
\maketitle

\begin{abstract}
In this paper, we evaluate Apache Spark for a data-intensive machine learning problem. Our use case focuses on policy diffusion detection across the state legislatures in the United States over time. Previous work on policy diffusion has been unable to make an all-pairs comparison between bills due to computational intensity. As a substitute, scholars have studied single topic areas.

We provide an implementation of this analysis workflow as a distributed text processing pipeline with Spark dataframes and Scala application programming interface.  We discuss the challenges and strategies of unstructured data processing, data formats for storage and efficient access, and graph processing at scale.
\end{abstract}


\section{Introduction}
\label{sec:intro}

Policy diffusion occurs when government decisions in a given jurisdiction are systematically influenced by prior policy choices made in other jurisdictions~\cite{diffusion}. While policy diffusion can manifest in a variety of forms, we focus on a type of policy diffusion that can be detected by examining similarity of legislative bill texts. Our dataset is based on the LexisNexis StateNet~\cite{lexis} and contains a total of more than 7 million legislative bills from 50 US states from 1991 to 2016. We aim to identify groups of legislative bills from different states falling into the same diffusion topic, to perform an all-pairs comparison between the bills within each topic, and to identify paths connecting specific legislative proposals on a graph. 

The causes and the extent to which policies spread across state legislatures is of substantive importance to political scientists, with implications for the states as laboratories of democracy~\cite{walker, gray, berry, balla, volden}. Previous work has been unable to make an all-pairs comparison between bills for a lengthy time period, as we do in this paper, because of computational intensity: a brute-force all-pairs calculation between the texts of the state bills yields $O(10^{13})$ distinct pairs. As a substitute, scholars have studied single topic areas~\cite{balla, Kreitzer, Mossberger, Pacheco, Shipan}, however, these areas are also the most likely to diffuse and thus tell us little about the extent to which law traverses across state borders.


\begin{figure}[h]
\centering
\includegraphics[width=.8\textwidth]{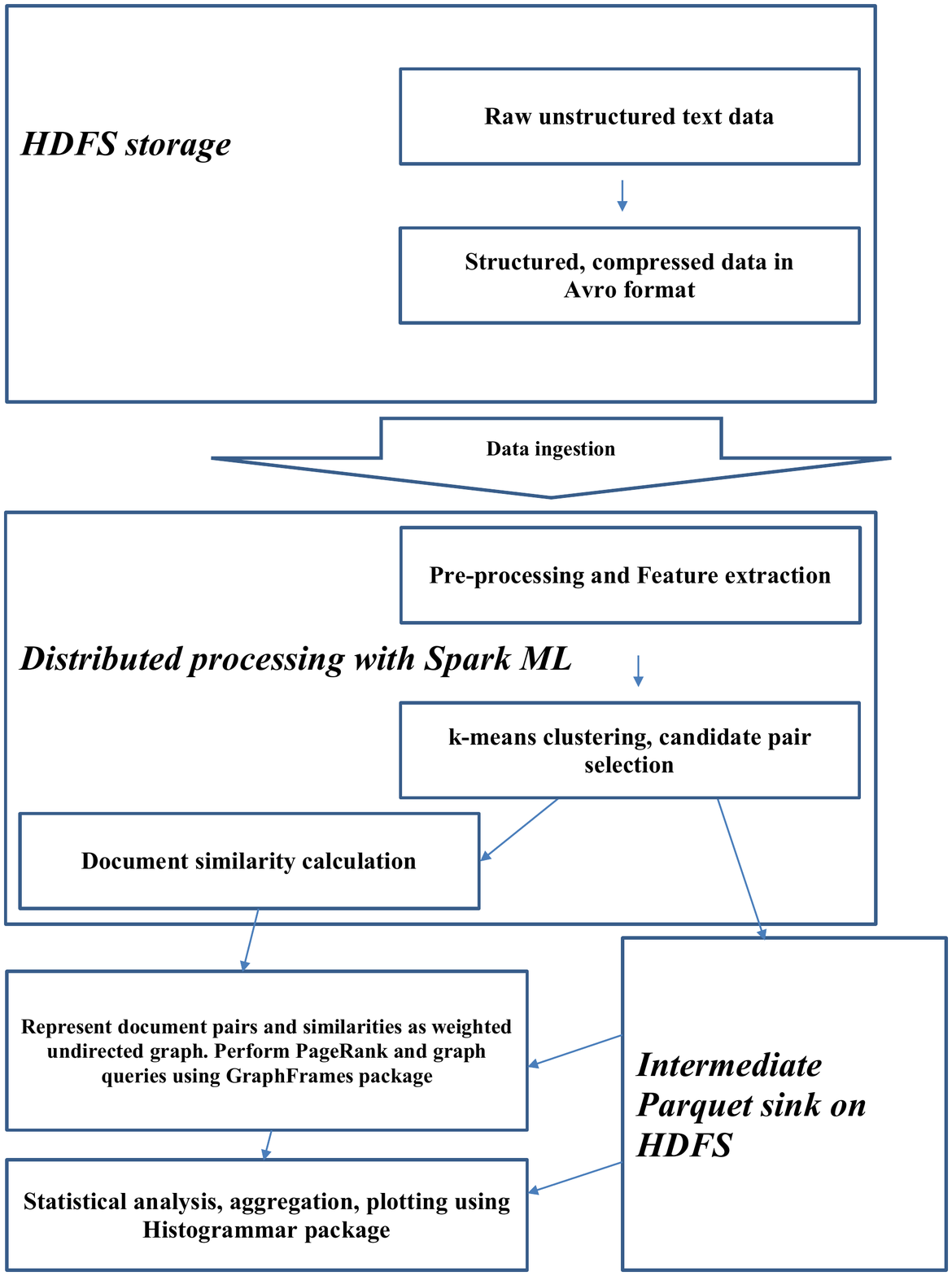}
\caption{The policy diffusion analysis pipeline.}
\label{fig:flowchart}
\end{figure}

Our analysis pipeline, which is summarized in Fig.~\ref{fig:flowchart}, consists of the following five stages: (1) data ingestion, (2) pre-processing and feature extraction, (3) candidate document selection, (4) document pair similarity calculation and (5) policy diffusion detection by ranking or on a graph. Apache Spark~\cite{rdds}, a high-performance distributed computing framework for large-scale data processing, is at the core of the pipeline implementation. Spark uses directed acyclic graph (DAG) instead of MapReduce execution engine, allowing to process multi-stage pipelines chained in one job. It is closely integrated with Apache Hadoop ecosystem and can run on YARN, use Hadoop file formats, and HDFS storage~\cite{yarn}. Spark provides an ability to cache large datasets in memory between stages of the calculation, allowing to reuse intermediate results of the computation in iterative algorithms like $k$-means clustering and improves fault tolerance by taking advantage of the data replication in HDFS and check-pointing.

The framework is applicable to a wider class of fundamental text mining problems of finding similar items, including plagiarism and mirror website detection~\cite{massivedat}. 

The implementation of the pipeline uses a mix of RDD based API along with the dataframes, allowing to take advantage of the Catalyst query optimizer and direct operations on serialized data, available in Spark 2.0.0~\cite{sparksql}. We use GraphFrames~\cite{graphframe} for dataframe-based graph algorithms calculation and graph queries. 

Scala is a high-level programming language for the Java virtual machine (JVM). We chose Scala over Python or R programming languages --- commonly considered a standard in text analysis research communities --- because the optimal performance of Spark is most likely to be achieved in that language. Spark code written in Python or R is often slower than equivalent code written in Scala, since it is statically typed and the cost of JVM communication from Python to Scala can be very high. 

Big data applications require a mix of processing techniques, data sources and storage formats. We chose Apache Avro row-based format and Apache Parquet columnar format for this pipeline instead of data formats like CSV and JSON, commonly used by scholars in this area, to take advantage of advanced compression and serialization, crucial for big data applications implemented with Spark. 

Interactive statistical analysis tools compatible with Spark have been implemented in the Histogrammar package~\cite{hg_soft}: a cross-platform suite of data aggregation primitives for making histograms, calculating descriptive statistics and plotting. Histogrammar allows to aggregate data using functional primitives, summarizing a large dataset with discretized distributions, using lambda functions and composition rather than a restrictive set of histogram types.


We observe stable execution of memory-intensive text processing jobs with large number of executor containers yielding efficiencies greater than 80\% for the largest dataset considered in the study. 

The paper is organized as follows: we start with the hardware specifications and details on the Spark cluster setup and Hadoop ecosystem; in the Section III the policy diffusion detection method and the core modules of the pipeline are described; Section IV introduces the Histogrammar tool for interactive data aggregation and plotting applied to the policy diffusion problem; finally, Section V summarizes the performance of the core modules of the pipeline and discusses the optimization. Section VI concludes the paper.


\section{Hardware specifications}
\label{sec:hardw}

An SGI Hadoop Linux cluster consisting of 6 worker nodes and 4 service 
nodes is used to deploy Apache Spark. The cluster is configured with all the servers mounted on one rack and interconnected using a 10 Gigabit Ethernet switch. Intel Xeon CPU E5-2680 v2 @ 2.80GHz CPU processors, with each worker node having 256 GB of memory and 10 TB of hard disk allow to achieve high performance and handle distributed workloads. 

The Hadoop cluster is configured without single points of failure by using two separate machines as name nodes. Spark applications are scheduled using YARN resource manager with dynamic resource allocation enabled. HDFS distributed file system is chosen to improve data locality by means of replication. 



\section{Text processing pipeline: core modules}
\label{sec:tfidf}

This section discusses the core modules of the text processing pipeline for policy diffusion detection.

\subsection{Data ingestion}

During ingestion step the raw unstructured data are converted into Apache Avro format having following schema:

\begin{lstlisting}
\begin{center}
{"namespace": "bills.avro",
 "type": "record",
 "name": "Bills",
 "fields": [
     {"name": "primary_key", "type": "string"},
     {"name": "content", "type": "string"}
     {"name": "year", "type": "int"},
     {"name": "state",  "type": "int"},
     {"name": "docversion", "type": "string"}
 ]   
}
\end{center}
\end{lstlisting}

where the $primary\_key$ field is a unique identifier of the elements in the dataset constructed from year, state and document version. The $year$, $state$ and $docversion$ fields are used to construct predicates and filter the data before the all-pairs similarity join calculation.  

The $content$ field stores the entire legislative proposal as a unicode string. It is only used for feature extraction step, and is not read into memory during candidate selection and filtering steps, thanks to the Avro schema evolution property.

Avro schema is stored in a file along with the data. Thus, if the program reading the data expects a different schema this can be easily resolved by setting the $avro.input.schema.key$ in the Spark application, since the schemas of Avro writer and reader are both present.


\subsection{Pre-processing and Feature extraction}

The feature extraction step consists of a sequence of Spark ML transformers intended to produce numerical feature vectors as a dataframe column. The resulting dataframe is fed to Spark ML $k$-means estimator, later used to calculate the all-pairs join, and subsequently during the graph analysis step with GraphFrames. 


\subsubsection{Data cleaning and stop word removal}

The raw text of legislative proposals from the StateNet dataset contains a lot of spurious white spaces and non-alphanumeric characters, which bare no meaning for analysis of legislative bills and often represent an obstacle for tokenization. The cleaner is implemented as a column-based user defined function (UDF). 

The words appearing very frequently in all the documents across the corpus (stop words) are excluded by means of \textit{StopWordsRemover} transformer from Spark ML, which takes a dataframe column of unicode strings and drops all the stop words from the input. The default list of stop words for English language is used in this study.

\subsubsection{Bag-of-words and the $N$-gram model}

In the bag-of-words model, text is represented as a multiset of words, disregarding grammar and word order but keeping multiplicity. The $N$-gram model, on the other hand, preserves the spatial information about the order within the multiset. Conceptually, the bag-of-word model can be viewed as a special case of the $N$-gram model with $N=1$. 

We use a regular expression based tokenizer which produces a dataframe column having an array of strings per row.
The \textit{NGram} transformer from Spark ML takes a sequence of strings from the output of tokenizer and converts it to a sequence of space-delimited strings of $N$ consecutive words, which are optionally added to the bag-of-word features to improve accuracy. 

\subsubsection{Term frequency and inverse document frequency calculation}
Term frequency-inverse document frequency (TF-IDF) is a feature vectorization method used to reflect the importance of a term to a document in the corpus. TF-IDF is implemented in two classes in Spark ML.






\textit{HashingTF} class implements a transformer, which takes tokenized documents and converts them into fixed-length feature vectors by means of the hashing trick. A raw feature is mapped to an index by applying the \textit{MurmurHash 3} hash function. The \textit{IDF} estimator is fit on feature vectors created from \textit{HashingTF}. It down-weights columns which appear frequently in the corpus. 

\subsubsection{Dimensionality reduction}

Singular value decomposition (SVD) is applied to the TF-IDF document-feature matrix to extract concepts which are most relevant for classification~\cite{lsa}.   

SVD factorizes the document-feature matrix $M$ ($m\times n$) into three matrices $U$,$\Sigma$ and $V$, such that: 
\begin{equation}
M =  U \cdot \Sigma \cdot V^{T},
\end{equation}
having $m \times k$, $k \times k$ and $k \times n$ dimensions correspondingly, where $m$ is the number of legislative bills ($O(10^6)$), $k$ is the  number of concepts, and $n$ is the number of features ($2^{14}$). Following inequalities hold:
\begin{equation}
m \gg n \gg k.
\end{equation}

The left singular matrix $U$ is represented as row-oriented distributed matrix while $\Sigma$ and $V$ matrices are sufficiently small to fit into the Spark driver memory. 

\subsection{Candidate selection and clustering}
\label{sec:cluster}

Focusing on the document vectors which are likely to be highly similar is essential for all-pairs comparison at scale.
Modern studies employ variations of nearest-neighbor search, locality sensitive hashing~\cite{massivedat}, as well as  sampling techniques to select a subset of rows of TF-IDF matrix based on the sparsity~\cite{dimsum}. Our approach utilizes $k$-means clustering to identify groups of documents which are likely to belong to the same diffusion topic, reducing the number of comparisons in the all-pairs similarity join calculation. 

The $features$ dataframe column is passed to the \textit{KMeans} estimator which generates \textit{KMeansModel} with a given number of cluster centroids. $k$-means clustering subdivides $N$ vectors in the feature space into $k$ clusters so that each vector belongs to a cluster with the nearest centroid, used to initialize the cluster. 

Given an initial set of $k$ cluster centroids $m_{i}^{(0)}$, where $i = 0...N$ the algorithm yields:


\begin{equation}
S_{i}^{(t)}={\big \{}x_{p}:{\big \|}x_{p}-m_{i}^{(t)}{\big \|}^{2}\leq {\big \|}x_{p}-m_{j}^{(t)}{\big \|}^{2}\ ,\forall j,1\leq j\leq k{\big \}},
\end{equation}
where each $x_{p}$ is assigned to exactly one $S^{(t)}$ during the iteration $t$.

During the update step, the means of the clusters are assigned to be the new clusters centroids on the next iteration:
\begin{equation}
m_{i}^{(t+1)}={\frac {1}{|S_{i}^{(t)}|}}\sum _{x_{j}\in S_{i}^{(t)}}x_{j}
\end{equation}
and the procedure is repeated before the convergence based on the within-cluster sum of squares (WCSS) objective is reached.


The optimum number of clusters has been determined empirically, by repeating the calculating for a range of values of $k$ and scoring them on a processing time versus WCSS plane. While processing time has been increasing with $k$, the WCSS gain has appeared to slow down significantly in the neighborhood of $k \approx 150$ for a 3 state subset and $k \approx 400$ for the entire dataset.


\subsubsection{Number of permutations}
Requesting algorithm to focus on the combinatorial pairs belonging to the same clusters reduces the number of comparisons in the all-pairs similarity join by 2-3 orders of magnitude, keeping the bill pairs belonging to the same diffusion topics with high probability.  

Indeed, starting with a total of $N=212768$ legislative proposals in a 3 state subset of the dataset, we would get a total of: $N\cdot(N-1)/2 = 2.26\times 10^{10}$
distinct combinatorial pairs to compare. Considering $k = 150$ classes for $k$-means clustering and assuming a uniform distribution of samples among these clusters we would get: $M = N/k = 212768/150 \approx 1418$, resulting in $M\cdot(M-1)/2\cdot k = 1418\cdot(1418-1)/2\cdot150 \approx 1.5\times 10^8$ combinatorial pairs, which is roughly 2 orders of magnitude less compared to the case with no clustering. The actual distribution among $k$-means clusters for this sample has shown a mean occupancy of $1467.3$ documents per cluster, with the standard deviation of $9562.4$, and the maximum occupancy of $110794$ documents per cluster, yielding a good reduction in the number of pairwise comparisons.

\subsection{Document similarity calculation}

The $k$-means clustering algorithm assigns vectors in the feature space to clusters by 
minimizing the WCSS objective. The step after that -- all-pairs document similarity calculation -- is performed within each cluster. Cosine, Jaccard, manhattan and Hamming similarity measures are considered.

The \textit{SimilarityMeasure} trait provides a common interface for similarity calculation between feature vectors reaching into Spark's private linear algebra code to use \textit{BLAS} dot product. Each similarity measure is implemented as an object extending the \textit{SimilarityMeasure} class and each implementing its own $compute$ method for dot product.







 



We convert Cosine, manhattan and Hamming distances to similarities assuming inverse proportionality, and re-scale all similarities to a common range:
\begin{equation}
S_{M} = \frac{100}{1 + D_{M}}
\end{equation}
an extra additive term in the denominator serves as a regularization parameter for the case of identical vectors.






\subsection{Policy diffusion detection}

The policy diffusion detection tool can be used in a number of modes: 
\begin{itemize}
\item identification of groups of diffused bills in the dataset given a diffusion topic, 
\item discovery of diffusion topics, 
\item identification of minimum cost paths connecting two specific legislative proposals on a graph, and, possibly, 
\item identification of the most influential US states for policy diffusion.
\end{itemize}

We use supervised pre-training on a set of diffusion topics labeled by an expert to tune the classification algorithm to achieve a better accuracy.

Below is an example output of the classifier for the test performed on a subset of legislative proposals having "Stand your ground" diffusion topic.

\textbf{Input:}
A set of bills on the topic: FL/2005/SB436, MI/2005/HB5153, MI/2005/HB5143, SC/2005/HB4301, MI/2005/SB1046, and a probe bill: FL/2005/SB436.

\textbf{Output:}
A set of top similarity bills from the test set contained all the samples labeled as having "Stand your ground" diffusion topic by an expert:
\begin{figure}
\centering
\begin{BVerbatim}
FL/2005/SB436, MI/2005/SB1046: 91.38,
FL/2005/SB436, MI/2005/HB5143: 91.29,
FL/2005/SB436, MI/2005/HB5153: 91.18,
FL/2005/SB436, SC/2005/SB1131: 82.89,
FL/2005/SB436, SC/2005/HB4301: 81.86,
FL/2005/SB436, SC/2011/SB1415: 77.11.
\end{BVerbatim}
\end{figure}

\subsection{Reformulating the problem as a network (graph) problem}



Some policy diffusion questions are easier answered if the problem is formulated as a graph analysis problem. The dataframe output of the document similarity step is mapped onto a weighted undirected graph, considering each unique legislative proposal as a node and a presence of a document with similarity above a certain threshold as an edge with a weight attribute equal to the similarity. The PageRank and Dijkstra minimum cost path algorithms are applied to detect events of policy diffusion and the most influential states.

A GraphFrame is constructed using two dataframes (a dataframe of nodes and an edge dataframe), allowing to easily integrate the graph processing step into the pipeline along with Spark ML, without a need to move the results of previous steps manually and feeding them to the graph processing module from an intermediate sink, like with isolated graph analysis systems. 

\section{Interactive analysis}


This section describes the tools and techniques used for interactive part of the analysis in read-eval-print loop (REPL) shell.

Histogrammar~\cite{hg_soft} is a suite of data aggregation primitives for making histograms, calculating descriptive statistics and plotting. A few composable functions can generate many different types of plots, and these functions are reimplemented in multiple languages and serialized to JSON for cross-platform compatibility. Histogrammar allows to aggregate data using cross-platform, functional primitives, summarizing a large dataset with discretized distributions, using lambda functions and composition rather than a restrictive set of histogram types.

Histogrammar primitives are order-independent commutative monoids designed for distributed computing and cross-platform compatibility. As a data analyst, you just express your data aggregation in terms of nested Histogrammar primitives and pass it to any system for evaluation. Since all of the logic of what to fill is encoded in your lambda functions, the aggregation phase is automatic.

Moving the logic of data analysis out of the for loop allows the analyst to describe an entire analysis declaratively. A whole analysis can be wrapped up in subdirectories like

\begin{lstlisting}[style=myScalastyle]
Label(
    dir1 = Label(
        hist1 = Bin(...),
        hist2 = Bin(...)),
    dir2 = ...)
\end{lstlisting}
    
This tree gets filled the same way as a single histogram, because the Label collection is a primitive just like Bin.

Thus, analysis code can be independent of where the data are analyzed. This is especially helpful for aggregating data in hard to reach places: across a distributed system like Apache Spark, on a GPU coprocessor, or through a thin bandwidth connection.

Histogrammar has front-end extensions to pass its aggregated data to many different plotting libraries, including Bokeh and Matplotlib.

Histogrammar also has back-end extensions for aggregating data from different frameworks. It can therefore be thought of as a common language for aggregating and then plotting data, so that every plotting library doesn't have to have individual hooks for every kind of aggregation system. 

An example interactive analysis in spark-shell REPL is provided in Appendix A.

\section{Performance evaluation}
\label{sec:perf}


This section discusses the core algorithms, types of transformations used in the analysis, partitioning, check-pointing and shuffle among the stages of calculation. 

The policy diffusion analysis involves $\bf{map}$, $\bf{filter}$, $\bf{join}$ and $\bf{aggregateByKey}$ transformations. 
All-pairs similarity calculation involves a two-sided $\bf{join}$ transformation with wide dependencies among partitions, resulting in an order of 100 TB of intermediate data and intensive shuffles, making the analysis challenging. The cost of failure for a partition with wide dependencies is rather high, since it requires a number of partitions to be re-computed, especially if the lineage graph is rather long. An intermediate Parquet sink is introduced between the two main steps of the computation (separating the feature extraction and document classification steps) to break the RDD's lineage. 

\subsection{All-pairs similarity calculation}
\label{sec:core}

The most compute and shuffle intensive part of the pipeline is the all-pairs document similarity calculation. To scale the solution up to large dataset sizes an efficient candidate selection step via $k$-means clustering is introduced~(\ref{sec:cluster}).

Once all rows of the pre-processed dataset $D$ are subdivided into $k$ clusters, a copy of the clustered dataset $\tilde{D}$ is broadcasted to each partition across the nodes of Spark cluster.
All combinatorial pairs of primary keys $(pk_{j}, pk_{k})$ corresponding to the documents are calculated in each partition,
filtered by state and predicted cluster label. The result is then aggregated into an array of pairs of primary keys ${(pk_{j}, pk_{k})}$ and combined. The RDD checkpoint is introduced following this step.

Next, the two-sided join is performed to calculate the and all-pairs similarity as described in~\ref{algos}. 
\begin{figure}
\label{algos}
\begin{algorithmic}[1]
\Procedure{Join on the left key}{$\tilde{D}$}
\State for each $(pk_{j}, pk_{k})$ in RDD calculate $(pk_{j},(pk_{j}, pk_{k}))$
\State join with the dataset $\tilde{D}$ on the left key
\EndProcedure

\Procedure{Join on the right key}{$\tilde{D}$}
\State for each $(pk_{j}, pk_{k})$ and feature vector $v_{j}$ in RDD calculate $(pk_{k},((pk_{j}, pk_{k}),v_{j})$ 
\State join with the dataset $\tilde{D}$ on the right key
\EndProcedure

\Procedure{Calculate similarities}{$threshold$}
\State for each $(pk_{j}, pk_{k})$ and feature vector $(v_{j}, v_{k})$ in RDD calculate $((pk_{j}, pk_{k}),(v_{j}, v_{k}))$

\State for each $((pk_{j}, pk_{k}),(v_{j}, v_{k}))$ calculate similarity between $v_{j}$ and $v_{k}$: $((pk_{j}, pk_{k}),S_{jk})$
       Filter $S_{jk} >$ threshold
\EndProcedure
\end{algorithmic}
\caption{Two-sided join and all-pairs similarity calculation.}\label{euclid}
\end{figure}
The DAG visualization of two-sided join and all-pairs similarity calculation is provided in Appendix B.


\subsubsection{Shuffle and partitioning}

\begin{figure}[h!]
 \centering
  \includegraphics[width=0.6\textwidth]{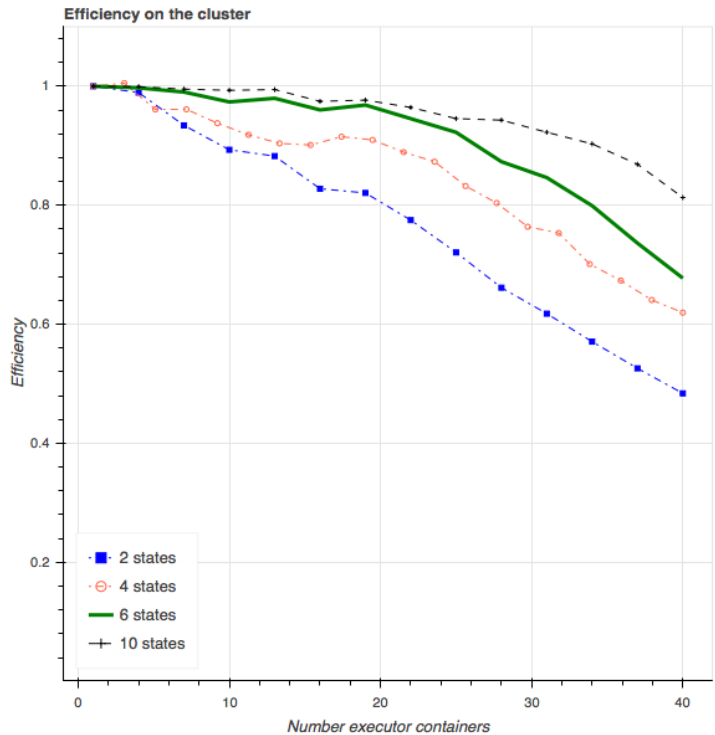}
 \caption{Efficiency as a function of the number of executor containers used in the calculation for different dataset sizes.}
 \label{fig:efficiency}
\end{figure}

Spark applications have been deployed on the cluster (\ref{sec:hardw}) with up to 40 executor containers each using 3 executor cores and 15 GB of RAM per JVM. We use external shuffle service inside the YARN node manager, observing an improved stability of memory-intensive jobs with larger number of executor containers.
Fig.~\ref{fig:efficiency} shows the efficiency of the computation on the Spark cluster, defined as:
\begin{equation}
E = \frac{T_{0}}{N_{exec}\cdot T_{N}}
\end{equation}
where $T_{0}$ is the processing time on a single executor and $T_{N}$ is the processing time using $N_{exec}$ executor containers. The total processing time is composed of the executor compute time, shuffle read-write time, task serialization and deserialization times, excluding scheduler delays. The efficiency is calculated for four distinct samples containing legislative proposals from 2, 4, 6 and 10 states respectively as a function of the number of executor containers used in the calculation.
As seen, the efficiency decreases down to 50\% for the case of 2 state sample, which is due to a relatively small problem size. The efficiency in the high-executor region is improved as the sample size increases, staying above 80\% for the 10 state sample. 

\begin{figure}[h!]
 \centering
  \includegraphics[width=0.6\textwidth]{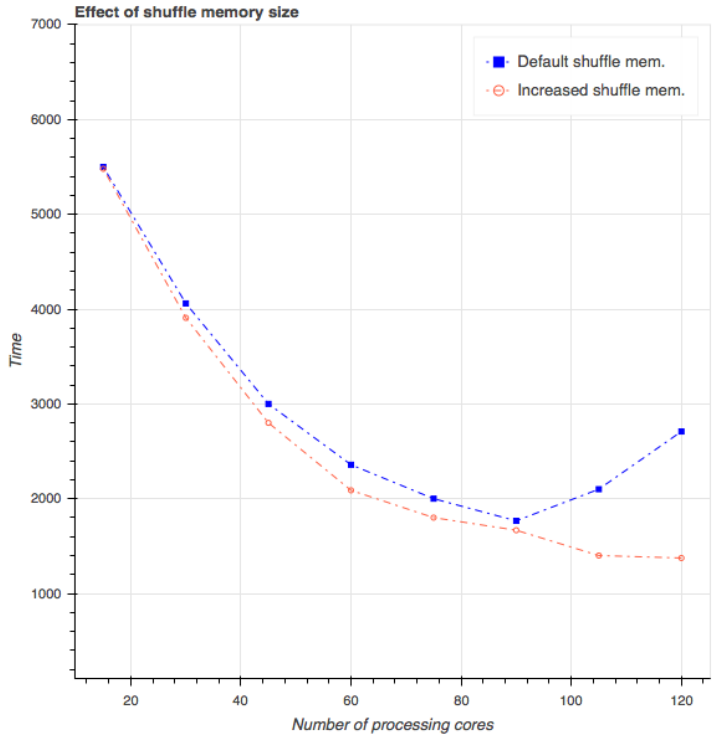}
 \caption{Processing time as a function of number of virtual cores in the Spark cluster. Comparison of two curves illustrate the effect of the shuffle memory spill fraction.}
 \label{fig:scaling2}
\end{figure}

An intensive shuffle across partitions of the dataset has been identified as the main cause of efficiency decrease. Fig.~\ref{fig:scaling2}, shows the scaling of the document similarity calculation step as a function of number of processing cores, as well as the effects of changing the fraction of Java heap space used during shuffles. If the specified threshold for in-memory maps used for shuffles is exceeded, the contents will begin to spill to disk.
Increasing the value of the memory fraction to 50\% of the executor memory allowed to maintain a good scaling beyond 90 processing cores.

\textit{Partition} is a unit of parallelism in Spark. Proper partitioning is necessary to speed-up the computation and avoid out-of-memory errors. The algorithm described in section (\ref{sec:core}) avoids grouping by key, thus minimizing the shuffle and eliminating "struggler tasks" which arise due to an uneven distribution of documents over key-groups.



Let $S$ be a feature vector for a short document and $L$ be a feature
vector for a long document, i.e., $|S| \le |L|$.  The usual Jaccard
measure is
\begin{equation}
 J \ = \  \frac{|S|+|L|-|S-L|}{|S| + |L| + |S - L|}
\end{equation}
Now, define $\alpha$ such that $|S| = \alpha |T|$ where
$0 < \alpha \le 1$.  We consider the following weighted Jaccard
measure,
\begin{equation}
  J_w \ = \  \frac{|S|+|L|-w|S-L|}{|S| + |L| + w|S - L|}
\end{equation}
Suppose that the fraction of $S$ contained in $L$ is given by $r$.
Then, $|S-L|=(1+\alpha-2\alpha r)|L|$.  The weighted Jaccard distance
becomes,
\begin{equation}
  J_w \ = \
  \frac{(1+\alpha)|L|-w(1+\alpha-2\alpha r)|L|}{(1+\alpha)|L|+w(1+\alpha-2\alpha r)|L|}
\end{equation}
We would like to choose $w$ as a function of $\alpha$ such that $J_w$
is approximately $r$ for any $r$.
\begin{equation}
  w \ = \ \frac{(1-r)(1+\alpha)}{(1+r)(1+\alpha-2\alpha r)}
\end{equation}

\section{Conclusion}
\label{sec:conc}
The Apache Spark framework has been evaluated for the case of data-intensive machine learning problem. A text processing pipeline utilizing Avro serialization framework, Spark ML, GraphFrames, and Histogrammar suite of data aggregation primitives has been proposed in application to a policy diffusion problem. 

The proposed framework allows to efficiently calculate all-pairs comparison between legislative bills, estimate relationships between bills on a graph, and is potentially applicable to a wider class of fundamental text mining problems of finding similar items. 

Histogrammar tool, adopted as a part of the framework to enable interactive analysis, allows a researcher to perform analysis in Scala language, integrating well with Hadoop ecosystem.


\newpage
\section{Appendix A}
\label{sec:appa}

Histogrammar is available on Maven Central, a publicly accessible Java/Scala repository with dependency management.
To use it in Spark 2.0, one does not have to download anything. Following will start spark-shell and include the Histogrammar core as well as the packages needed to work with Spark SQL and Bokeh (\ref{lst:shell}).

\begin{lstlisting}[float=*,label={lst:shell}]
spark-shell --packages "org.diana-hep:histogrammar_2.11:1.0.3",
"org.diana-hep:histogrammar-bokeh_2.11:1.0.3",
"org.diana-hep:histogrammar-sparksql_2.11:1.0.3"
\end{lstlisting}

Following lines of Scala code in the interactive spark-shell produce an editable HTML file with the similarity distribution plot for all the bill pairs having at least one bill from Florida (\ref{lst:bokeh}).
\begin{lstlisting}[style=myScalastyle2,label={lst:bokeh}]
import org.dianahep.histogrammar._
import org.dianahep.histogrammar.bokeh._
import org.dianahep.histogrammar.sparksql._
import io.continuum.bokeh._

//UDF to filter data
def stateSelector_udf = udf((pk1: String,pk2: String) =>
{(pk1 contains "FL") || (pk2 contains "FL")})

//load and filter data
val data = spark.read.parquet("path").cache()
val filtered = data.filter(stateSelector_udf(col("pk1"),col("pk2")))

//create and fills the histogram
val hist = filtered.histogrammar(Bin(20, 0, 100, $"similarity"))

//plot the histogram and save
val plot = hist.bokeh(glyphType="histogram",glyphSize=3,fillColor=Color.Red)
                 .plot(xLabel="Similarity",yLabel="Num. pairs")
save(plot,"cosine_sim.html")
\end{lstlisting}

\begin{lstlisting}[style=myScalastyle2,label={lst:bokeh2}]
// Configure an ML pipeline 
val cleaner = new Cleaner()
      .setInputCol("content")
      .setOutputCol("cleaned")
      
val tokenizer = new RegexTokenizer()
      .setInputCol(cleaner.getOutputCol)
      .setOutputCol("words")
      .setPattern("\\W")

val remover = new StopWordsRemover()
      .setInputCol(tokenizer.getOutputCol)
      .setOutputCol("filtered")

val ngram = new NGram()
      .setN(nGramGranularity)
      .setInputCol(remover.getOutputCol)
      .setOutputCol("ngram")

val hashingTF = new HashingTF()
      .setInputCol(ngram.getOutputCol)
      .setOutputCol("keys")
      .setNumFeatures(numTextFeatures)

var idf = new IDF()
      .setInputCol(hashingTF.getOutputCol)
      .setOutputCol("features")

val pipeline = new Pipeline()
      .setStages(Array(cleaner, tokenizer, 
                 remover, ngram, hashingTF, idf))

 // Fit the pipeline 
 val model = pipeline.fit(train)
\end{lstlisting}

\begin{figure}[h!]
 \centering
  \includegraphics[width=0.6\textwidth]{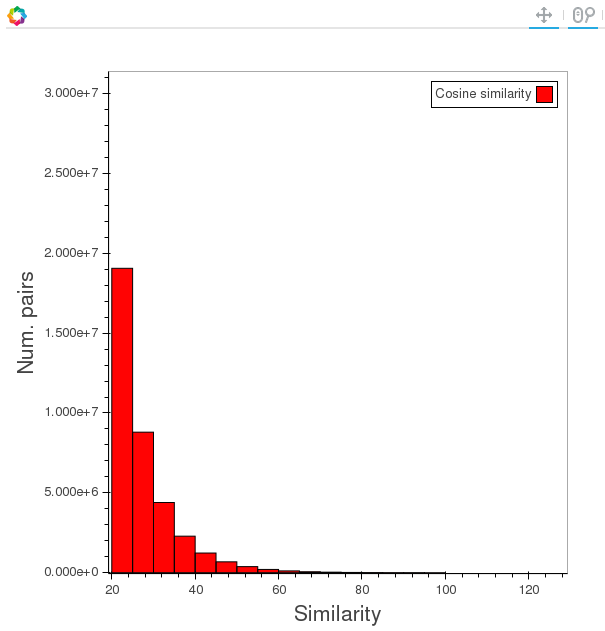}
 \caption{Similarity distribution produced with Histogrammar using Bokeh plotting front-end.}
 \label{fig:histo}
\end{figure}

\section{Appendix B}

\begin{figure*}[tb]
 \centering
  \includegraphics[width=0.99\textwidth]{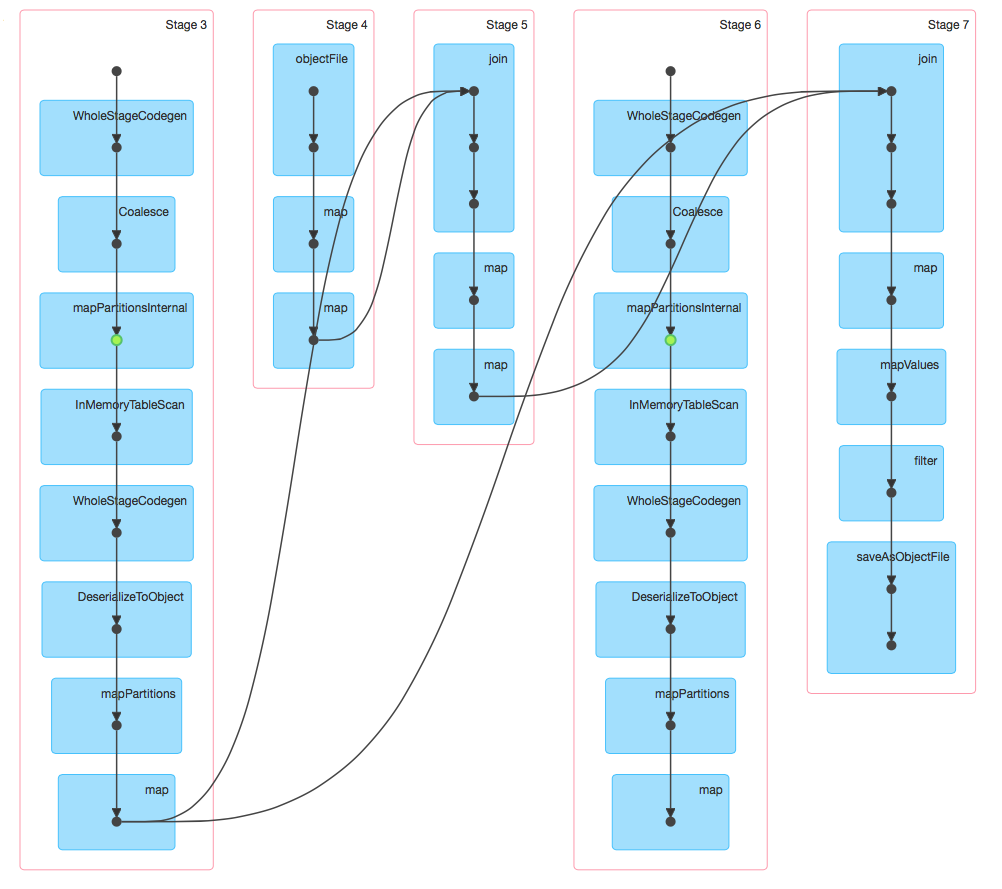}
 \caption{DAG visualization of two-sided join and all-pairs similarity calculation.}
 \label{fig:histo3}
\end{figure*}



\bibliographystyle{unsrt}  
\bibliography{references}  

\end{document}